%% file: main.tex
\def\year{2021}\relax
\documentclass[letterpaper]{article} 
\newcommand{\TODO}[1]{{{\textcolor{red}{#1}}}}
\usepackage{aaai21}  
\usepackage{times}  
\usepackage{helvet} 
\usepackage{courier}  
\usepackage[hyphens]{url}  
\usepackage{graphicx} 
\usepackage{amsfonts,amsmath,amssymb,amsthm}
\usepackage{bm,nicefrac}
\usepackage{natbib}  
\usepackage{caption} 
\usepackage{xcolor}
\usepackage{wrapfig}
\usepackage{verbatim}
\usepackage{courier}
\usepackage[switch]{lineno}
\usepackage{xspace}
\AtBeginDocument{}
\setkeys{Gin}{draft=false}      
\title{HOT-VAE: Learning High-Order Label Correlation for Multi-Label \\ Classification via Attention-Based Variational Autoencoders}

 \author {
         Wenting Zhao, \textsuperscript{\rm 1}
         Shufeng Kong, \textsuperscript{\rm 1}
         Junwen Bai, \textsuperscript{\rm 1}
         Daniel Fink, \textsuperscript{\rm 2}
         Carla Gomes \textsuperscript{\rm 1} \\
 }
 \affiliations {
     \textsuperscript{\rm 1} Department of Computer Science, Cornell University, USA \\
     \textsuperscript{\rm 2} Cornell Lab of Ornithology, Ithaca, NY, USA \\
     wzhao@cs.cornell.edu, \{sk2299, jb2467, daniel.fink\}@cornell.edu, gomes@cs.cornell.edu
 }
\begin{document}

\maketitle

\input{abstract}
\input{introduction}
\input{related}
\input{background}

\input{method}

\input{experiment}

\input{conclusion}

\section*{Acknowledgments}
We thank the anonymous reviewers for their valuable comments.
We thank Di Chen and Yiwei Bai for the helpful discussions.
We thank the eBird participants for their contributions and the eBird team for their support.
Daniel Fink was funded by The Leon Levy Foundation, The Wolf Creek Foundation, and the National Science Foundation (ABI sustaining: DBI-1939187).
The other authors were supported by NSF awards CCF-1522054 (Expeditions in computing) and CNS-1059284 (Infrastructure), AFOSR Multidisciplinary University Research Initiatives (MURI) Program FA9550-18-1-0136, ARO award W911NF-17-1-0187, and an award from the Toyota Research Institute.

\bibliography{references}

\newpage

\clearpage

\input{supp}

\end{document}

%% file: abstract.tex
\begin{abstract}
    Understanding how environmental characteristics affect biodiversity patterns, from individual species to communities of species, is critical for mitigating effects of global change.
    A central goal for conservation planning and monitoring is the ability to accurately predict the occurrence of species communities and how these communities change over space and time.
    This in turn leads to a challenging and long-standing problem in the field of computer science - how to perform accurate multi-label classification with hundreds of labels?
    The key challenge of this problem is its exponential-sized output space with regards to the number of labels to be predicted.
    Therefore, it is essential to facilitate the learning process by exploiting correlations (or dependency) among labels.
    Previous methods mostly focus on modelling the correlation on label pairs; however, complex relations between real-world objects often go beyond second order.
    In this paper, we propose a novel framework for multi-label classification, High-order Tie-in Variational Autoencoder (HOT-VAE), which performs adaptive high-order label correlation learning.
    We experimentally verify that our model outperforms the existing state-of-the-art approaches on a bird distribution dataset on both conventional F1 scores and a variety of ecological metrics.
    To show our method is general, we also perform empirical analysis on seven other public real-world datasets in several application domains, and Hot-VAE exhibits superior performance to previous methods.
\end{abstract}

%% file: introduction.tex
\section{Introduction}

The study of multi-label classification (MLC) is an active research area and has been receiving increasing attention in the past few decades; unlike traditional single-output learning, it is a task of predicting the presence and absence of multiple entities simultaneously given a sample with a set of features.
It finds applications in a wide range of domains including image recognition, natural language processing, and bioinformatics~\cite{8892612}.

One important field that is in urgent need for a scalable and accurate MLC approach is ecology.
The ability to accurately predict which species assemble into communities based on local environmental features is essential to understand how changes in the environment can be expected to impact biodiversity and to plan for the restoration and recovery of species communities in the face of environmental change~\cite{d2017spatial}.
This problem is represented as joint species distribution modelling (JSDM), which predicts species occurrences given environmental features and species interactions.
There are two key challenges in JSDM.
First, species communities are often comprised of very large numbers of individual species, presenting the challenge of learning complex high-dimensional interactions.
For example, bird communities are often comprised hundreds of individual species within a single region.
To consider how all subsets of all bird species interact is computationally intractable, thus being selective in how we model these interactions is a necessity.
On one hand, accounting for more interactions increases learning capacity but will be more computationally demanding and could have a higher risk of over-fitting.
On the other hand, one can focus on less or no interactions; however, many communities of species are known to include complex interactions among large numbers of species, so having oversimplifying assumptions leads to inaccurate predictions.
The second challenge for modeling the joint distribution of many species is the fact that the relationships between species change as important features of the environment change over space and time.
Thus, it is critical to understand how environment changes like climate change impact species interactions and the resulting structure of species communities~\cite{evans}.

\textbf{Our contributions:} We propose High-order Tie-in Variational Autoencoder (HOT-VAE), an attentioned-based VAE that leverages latent embedding learning and neural message passing to perform high-order label correlation learning and produce accurate multi-label predictions. More specifically:
(1) We introduce a two-branch VAE-based model with a replaceable, domain-specific encoder (i.e., one can choose an encoder that efficiently extracts feature information given an application domain) and a shared message passing neural network (MPNN) decoder where label correlations are computed.
(2) HOT-VAE is able to learn high-order label correlation using multiple-step message passing. It also produces label correlation conditioned on features. In other words, HOT-VAE can not only model the interactions between many species simultaneously, it can  also adapt the correlation between species to changing environmental factors.
(3) With a graph structure to reason about label correlation, we can easily incorporate prior knowledge, resulting in better empirical results.
(4) We perform thorough experimental evaluations on a JSDM dataset and seven other real-world datasets on a variety of metrics, and we show HOT-VAE outperforms (or is comparable to) other state-of-the-art MLC methods. On the JSDM dataset, we further evaluate HOT-VAE on several ecological metrics; the result suggests that HOT-VAE produces a meaningful improvement in the field of ecology.

%% file: related.tex
\section{Related Work}


We discuss three groups of MLC methods and how they are related to our approach.
The first group is binary relevance (BR) methods which treat a MLC problem as a number of independent binary classification problems~\cite{boutell2004learning,zhang2007ml}.
To add label correlation, probabilistic classifier chains (PCCs) stack binary classifiers sequentially and output one label at a time conditioned on all previously predicted labels~\cite{read2008multi,cheng2010bayes}.
Followup works extend PCCs to recurrent neural networks~\cite{wang2016cnn,nam2017maximizing}.
This group suffers from two issues: the quality of predictions can be highly dependent on label ordering, and the nature of autoregressive models prohibits them from parallel computation.

The second group deals with latent embedding, in which they learn one shared latent space representing both input features and output labels~\cite{bhatia2015sparse,yeh2017learning,tang2018multi,chen2019two}.
Most recently, \citet{bai2020disentangled} propose MPVAE: it learns VAE-based probabilistic latent spaces for both labels and features and aligns the latent representations using the Kullback–Leibler divergence.
These methods impose label-aware structure on the feature latent space, which is empirically showed to produce better predictive performance.
However, they only consider up to second-order label correlation: MPVAE's decoder is an multivariate probit model (MVP)~\cite{chen2018end} which employs a covariance matrix on labels to capture pairwise relations, and \citet{bhatia2015sparse} find k-nearest neighbors for label embeddings.
Another limitation is that these label correlations are global: when features change, the label interactions remain the same.
It is also not clear how they can incorporate prior knowledge on label structures.

The third group models label dependencies using graphical model representations~\cite{lafferty2001conditional,chen2019multi}.
Methods within this group often build a label graph, in which a node corresponds to a label, and edges represent how two labels interact with each other.
Most recently, \citet{lanchantin2019neural} propose LaMP, where they apply the attention mechanism from Transformer~\cite{vaswani2017attention} to learn how other labels contribute to the presence/absence of a label.
Further, they use an MPNN, a generalization of graph neural networks (GNN)~\cite{scarselli2008graph}, to pass messages among label nodes weighted by attention, thus modelling a conditional joint representation of output labels.
Compared to PCC methods, LaMP provides a scalable and flexible module to model label correlations that requires no label ordering and allows parallel computation, and its graphical structure making it a natural fit to impose constraints on labels.
However, LaMP follows a encoder-to-decoder architecture, and learning feature embeddings within this architecture has not yet been optimized.
It does not incorporate any label information in the feature embedding, which has been shown to be beneficial for making accurate predictions~\cite{yeh2017learning}.

This work leverages both the state-of-the-art latent embedding learning and powerful attention-based MPNNs to provide an accurate and scalable multi-label classifier, and we further extend the message passing module to model high-order label correlations to improve performance.


%% file: background.tex
\section{Background}
Let $\mathcal D$ denote the dataset $\{(\bm{x}_i,\bm{y}_i)\}_{i=1}^N$, where $\bm{x}_i \in \mathbb R^S$ is an input and $\bm{y}_i \in \{0,1\}^L$ is an output associated with sample $i$.
Input $\bm{x}_i$ can alternatively be an ordered set of $S$ elements, and output $\bm{y}_i$ has $L$ binary labels with 1 indicating the presence and 0 indicating the absence of a label.


\subsection{Aligned Variational Autoencoders}
A variational autoencoder (VAE) is a generative model which consists of an encoder, a decoder, and a loss function.
The encoder, denoted by $q_\theta(\bm{z}_i|\bm{x_i})$, is a neural network that maps features of samples $\bm{x_i}$ into hidden representations $\bm{z}$, which have a much lower dimensionality than that of $\bm{x_i}$.
$\bm{z}_i$ represents a multivariate Gaussian probability density, and by sampling from this distribution we obtain noisy values of $\bm{z}_i$.
The decoder, denoted by $p_\phi(\bm{x_i}|\bm{z}_i)$, is another neural network that reconstructs $\bm{z}$ to $\bm{x_i}$.
The loss function to be minimized is $\mathbb E_{\bm{z}\sim q_\phi}[\log p_\theta(\bm{x_i}|\bm{z})]-\mathcal K [q_\phi(\bm{z}|\bm{x_i})||P(\bm{z})]$, where
$\mathcal K$ is the Kullback–Leibler (KL) divergence, and
$P(\bm{z})$ is the prior which is a standard multivariate normal distribution.
The first term encourages the reconstruction of $\bm{x_i}$, and the second term penalizes the KL divergence between the approximated distribution $q_\theta(\bm{z}|\bm{x_i})$ and the prior $P(\bm{z})$ which imposes structure on the latent space.

VAE can be used for the task of classification if the decoder is to predict target $y_i$ instead of reconstructing feature $x_i$. Therefore, we rewrite the decoder and loss function as $p_\phi(\bm{y_i}|\bm{z_x})$ and $\mathbb E_{\bm{z_x}\sim q_\phi}[\log p_\theta(\bm{y_i}|\bm{z_x})]-\mathcal K [q_\phi(\bm{z_x}|\bm{x_i})||P(\bm{z_x})]$, respectively. In this case, it is desired that the prior $P(z_x)$ imposes domain-specific structure on the latent space rather than just imposing the conventional standard multivariate normal structure. Recently, MPVAE was proposed to use another VAE to learn a latent multivariate Gaussian distribution $q_\psi(\bm{z}_{\bm{y}}|\bm{y}_i)$, and align $q_\psi(\bm{z}_{\bm{y}}|\bm{y}_i)$ and $q_\phi(\bm{z}_{\bm{x}}|\bm{x}_i)$ by penalizing their KL divergence \cite{bai2020disentangled}. The feature and label VAEs are designed to share the same decoder. Thus, the label decoder is $p_\theta(y_i|z_y)$, and the loss function can be revised as:

\begin{align*}
    {KL}=&\frac{1}{2}(\mathbb E_{\bm{z}_{\bm{y}} \sim q_\phi}[\log p_\theta(\bm{y}_i|\bm{z}_{\bm{y}})]+\mathbb E_{\bm{z}_{\bm{x}}\sim q_\psi}[\log p_\theta(\bm{y}_i|\bm{z}_{\bm{x}})]) \\
    &-\beta\mathcal K [q_\psi(\bm{z}_{\bm{y}}|\bm{y}_i)||q_\phi(\bm{z}_{\bm{x}}|\bm{x}_i)],
\end{align*}

where $\beta$ is a hyper-parameter to control the similarity between the two latent Gaussian distributions.

\subsection{Attention Models}

\emph{Self-attention}, sometimes called \emph{intra-attention}, is a mechanism that assigns different importance to different positions of a sequence in order to focus on more important parts.
Self-attention has been used successfully in a variety of tasks such as natural language processing (NLP) and MLC.
The Transformer and the Generative Pre-trained Transformer 3 (GPT-3) \cite{brown2020language} are two well-known attention models in NLP achieving state-of-the-art performance.
The recent Label Message Passing (LaMP)~\cite{lanchantin2019neural} can be regarded as an application of the Transformer/GPT-3 on multi-label classification.

LaMP alternates between self-attention and feed forward layers. In the $t$-th self-attention layer, each label is represented by a node $v_i^t \in \mathbb R^d$. The attention weight $a_{ij}^t$ for a node pair $(\bm{v}_i, \bm{v}_j)$ is computed as:
\begin{align}
    \label{eq:alpha}
    e^t_{ij} = a(\bm{v}^t_i,\bm{v}^t_j) = \frac{(\mathbf{W}^q\bm{v}^t_i)^{\top}
    (\mathbf{W}^u\bm{v}^t_j)}{\sqrt{\smash[b]d}} \\
    \alpha^t_{ij} = \textrm{softmax}_j(e^t_{ij}) =  \frac{\textrm{exp}(e^t_{ij})}{\sum_{k \in \mathcal{N}(i)}{\textrm{exp}(e^t_{ik})}},
\end{align}
where $a(\cdot)$ is a dot product with node-wise linear transformations $\mathbf{W}^q \in \mathbb R^{d \times d}$ on node $\bm{v}_i^t$ and $\mathbf{W}^u \in \mathbb R^{d \times d}$ on node $\bm{v}_j^t$, scaled by $\sqrt{d}$;  $e_{ij}^t$ represents the raw importance of label $j$ to label $i$ and is further normalized by a softmax function to obtain $\alpha^t_{ij}$. Then, the attention $m_i^t$ message of $v_i^t$ is generated as:
\begin{gather}
    \label{eq:m_attention}
    M_{\textrm{atn}}(\bm{v}^t_i,\bm{v}^t_j) = \alpha^t_{ij} \mathbf{W}^v \bm{v}^t_j,\\
    \bm{m}^t_i = \bm{v}^t_i + 
    \sum_{j\in \mathcal{N}(i)} M_{\textrm{atn}}(\bm{v}^t_i,\bm{v}^t_j), \label{eq:mt}
\end{gather}
where $\mathbf{W}^v \in \mathbb R^{d \times d}$ is a node-wise linear transformation.

After going through the $t$-th feed forward layer $U^t$, we obtain $v_i^{t+1}$ in the $(t+1)$-th self-attention layer as:

\begin{align}
    {\bm{v}}^{t+1}_i &= \bm{m}^t_i + U^{t}(\bm{m}^{t}_i; \bm{W}).
    \label{eq:u_attention}
\end{align}

%% file: method.tex
\section{Method: Attention-based VAE \\ for High-order Correlation}

\begin{figure*}
  \centering
  \includegraphics[width=0.45\textwidth]{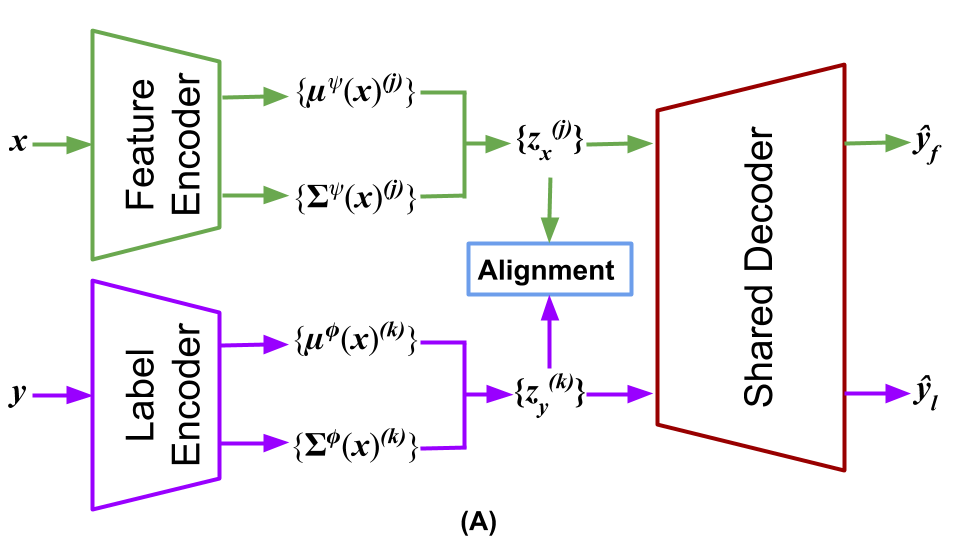}
  \includegraphics[width=0.45\textwidth]{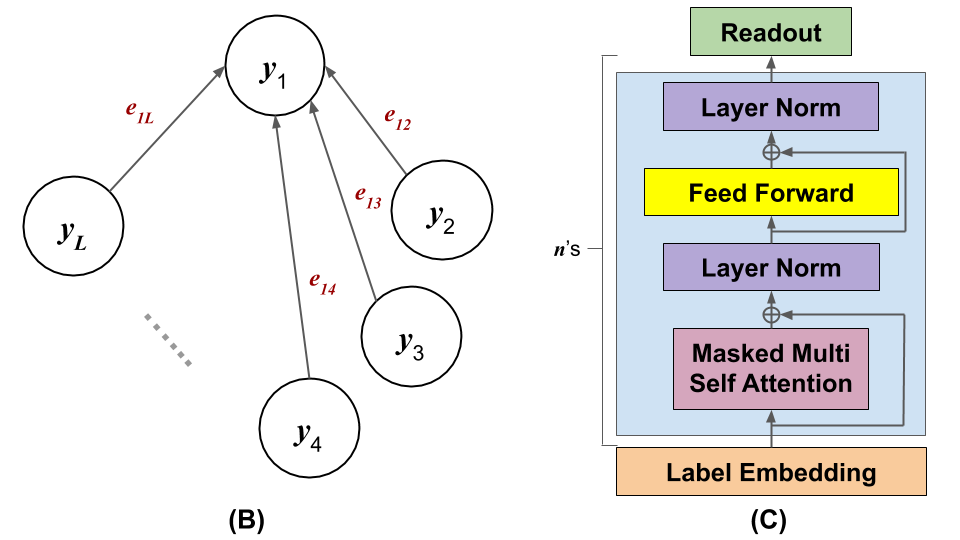}
  \caption{Model architecture of HOT-VAE. (A): Overall network architecture. The feature encoder maps features $x$ to a set of probabilistic latent subspaces using a neural network parameterized by $\psi$. Similarly, the label encoder with parameter $\phi$ maps labels $y$ to another set of probabilistic latent subspaces. Then, sampled from their own distributions, $\{z_{x}^{(j)}\}$ and $\{z_{y}^{(k)}\}$ are separately fed into a shared decoder. Finally, for the feature branch, the decoder outputs label prediction $\hat{y}_f$, and for the label branch, the decoder outputs reconstruction $\hat{y}_l$. At inference, only the feature branch is used. (B): The shared decoder is a graph with each node being a label. An edge is connecting two nodes if we believe correlation exists between them. By default, there is an edge between every pair of labels. The figure shows all the other nodes are sending messages to the $y_1$ node with learned attention weights $e$. (C): Decoder layers. At each layer, masked multi-head self-attention is computed and goes through a feed-forward operation. $n$ is the number of layers. We note that these layers are shared by all label nodes.}
  \label{fig:model}
\end{figure*}

We propose HOT-VAE, a novel two-branch variational autoencoder model building on top of attention-based neural message passing networks for MLC, which can learn feature embeddings representing both input features and output labels, perform high-order correlation learning, and flexibly incorporate prior knowledge on label structures.
The illustration of the framework is shown in Figure~\ref{fig:model}.
At training, the feature encoder and the label encoder first map features and labels to a set of Gaussian subspaces respectively.
There are many possible ways to parameterize the encoders; common choices are multi-layer perceptrons and graph networks such as graph convolutional networks and message passing neural networks.
Then, the shared decoder makes a prediction on labels based on the samples from the feature Gaussian subspace and recovers the input labels based on the samples from the label Gaussian subspace.
There are two message passing modules.
One passes attention from features to labels, and the other passes attention from labels to labels.

\subsection{Learning and Aligning Probabilistic Subspaces}
We assume that both the feature embedding and the label embedding have $d$ dimensions.
If each encoder only outputs one Gaussian subspace, in the case of it being an MLP, $\mathcal D[q_\phi(\textbf{z}|\textbf{y})||q_\psi(\textbf{z}|\textbf{x})]$ is simply the KL divergence between two multivariate normal distributions.
Since both distributions have diagonal covariance matrices, we can derive the KL divergence to be the following:
\begin{small}
\begin{equation}\label{eq:kl}
\begin{split}
    \mathcal{L}_{\text{KL}}(\bm{x},\bm{y})=&\beta[\sum_{i=1}^d\log\frac{\Sigma^\psi_{i,i}(\bm{x})}{\Sigma^\phi_{i,i}(\bm{y})}-d+\sum_{i=1}^d\frac{\Sigma^\phi_{i,i}(\bm{y})}{\Sigma^\psi_{i,i}(\bm{x})}+\\
    &\sum_{i=1}^d\frac{(\mu^\psi_i(\bm{x})-\mu^\phi_i(\bm{y}))^2}{\Sigma^\psi_{i,i}(\bm{x})}]
\end{split}
\end{equation}
\end{small}
However, there are features which MLPs cannot encode, such as English texts and graphs.
To have a general multi-label classifier, encoders need to be flexible with regards to network architecture to deal with different types of inputs.
Then, it becomes possible that the feature encoder and the label encoder output different numbers of Gaussian subspaces.
For example, if we adopt the Transformer encoder~\cite{vaswani2017attention} to be the feature encoder and the label encoder, the two encoders may generate two sets of Gaussian subspaces of varying sizes.
This is because for every dimension of the input, a mean and a variance are computed, and features and labels often differ in their dimensionality.
For example, let us consider when the input text is seven words - ``my favorite football player lost a game'' and the labels are ``sad'' and ``angry''.
The feature transformer outputs one Gaussian subspace for each of the seven words, and the label transformer outputs two subspaces.
To overcome this issue, suppose the feature encoder outputs $J$ subspaces and the label encoder outputs $K$ subspaces, we compress $J$ subspaces and $K$ subspaces into one subspace by computing a mean vector for $\mu$ and for $\Sigma$, respectively.
Formally, $\mu^\psi(\bm{x})$ and $\Sigma^\psi(\bm{x})$ now become:
\begin{align}
    \mu^\psi{(\bm{x})} &= \frac{1}{J}\sum_{j=0}^{J-1} \mu^\psi(\bm{x})^{(j)}\\
    \Sigma^\psi{(\bm{x})} &= \frac{1}{J}\sum_{j=0}^{J-1} \Sigma^\psi(\bm{x})^{(j)}
\end{align}
And we do the same for $\mu^\phi(\bm{y})$ and $\Sigma^\phi(\bm{y})$.
With these operations, we can again use Equation~\ref{eq:kl} to compute the divergence between two probabilistic latent spaces.
It is worth noting that because there is not a one-to-one relationship from one feature Gaussian subspace to one label Gaussian subspace, there is no point in aligning individual feature subspaces to individual label subspaces.

Lastly, although we collapse all Gaussian subspaces into a unified one to compute alignment, we still feed unmodified $\{\bm{z}^{(j)}_{\bm{x}}\}$ and $\{\bm{z}^{(k)}_{\bm{y}}\}$ into the shared decoder to keep as much information as possible.

\subsection{Learning High-Order Label Correlation}
We highlight three features of the shared decoder:
(1) The decoder computes the correlation between labels conditioned on features $\bm{x}$; thus, the label correlation becomes sensitive to changes in $\bm{x}$, which enables adaptive learning for label interactions.
(2) With the label graph, one can easily impose prior structure between labels by adding and removing edges. If it is known in advance two labels are independent from each other, then leaving out the edge connecting these two label nodes prevents the model from over-fitting and learning noise.
(3) Most importantly, the decoder is able to capture higher-order label correlation in a scalable way, as opposed to PCC methods~\cite{wang2016cnn,nam2017maximizing} where chain rules are used to model the joint probabilities of labels.

We now turn to a detailed description of HOT-VAE's decoder.
Labels are represented as embedded vectors $\{\bm{u}^t_1, \bm{u}^t_2, \dots, \bm{u}^t_L\}$, where $\bm{u}^t_i \in \mathbb R^d$ and initial $\bm{u}^{t=0}_i$ is obtained from a learnable embedding matrix $\textbf{W}^{y} \in \mathbb R^{L \times d}$.
Figure~\ref{fig:model}(C) shows an overview of the decoder layers.
Self-attention is computed based on Equations~\ref{eq:alpha}-\ref{eq:mt}, and feed forward is computed using Equation~\ref{eq:u_attention}.
Specifically, we use multi-head self-attention~\cite{vaswani2017attention}, so that a node can attend to multiple other nodes at once.
We also apply layer normalization~\cite{ba2016layer} around each of the attention and feedforward sublayers to alleviate training issues.
After the $n$'s layers (thus $\bm{u}^{t=0}_i$ becoming $\bm{u}^t_n$), a readout layer predicts each label $\hat{y}_i$, where a readout function $R$ projects $\bm{u}_i^{n}$ using a projection matrix $\textbf{W}^o \in \mathbb R^{d\times d}$.
The $i$th row of $\textbf{W}^o$ is denoted by $\textbf{W}^o_i$.
The resulting vector of size $L \times 1$ is then fed through an element-wise sigmoid function to produce the final probabilities of all labels:
\begin{equation}
   \label{eq:readout}
    \hat{y}_i = R({\bm{u}}^n_i;{\textbf{W}^o}) = \textrm{sigmoid}({\textbf{W}^o_i} {\bm{u}}^n_i).
\end{equation}

As mentioned above, to pass the message from encoder to decoder, 
we feed unmodified $\{\bm{z}^{(j)}_{\bm{x}}\}$ and $\{\bm{z}^{(k)}_{\bm{y}}\}$ into the shared decoder to keep as much information as possible.
For clarity, we look at the feature branch, and the label branch works similarly.
We denote the message passing module from $\{ {\bm{z}^{(j)}_{\bm{x}}} \}$ to label nodes by $\bm{W}_{\textrm{fy}}$ and the message passing module between label nodes by $\bm{W}_{\textrm{yy}}$.
We compute the initial state $\bm{u}^{t'}_i$ of the decoder w.r.t. the feature branch as:
\begin{gather}
    \bm{m}^t_i = \bm{u}^t_i + 
    \sum\limits_{j=1}^J M_{\textrm{atn}}(\bm{u}^t_i,\bm{z}^{(j)}_{\bm{x}}; \bm{W}_{\textrm{fy}}), \\
    \bm{u}^{t'}_i = \bm{m}^t_i + U_{\textrm{mlp}}(\bm{m}^{t}_i; \bm{W}_{\textrm{fy}}).
\end{gather}

After $\bm{u}^{t}_i$ is updated to $\bm{u}^{t^\prime}_i$ with feature information, the message passing between labels then begins,

\begin{gather}
    \label{eq:label_to_label_M}
    \bm{m}^{t'}_i = \bm{u}^{t'}_i + 
    \sum_{l \in \mathcal{N}(i)} M_{\textrm{atn}}(\bm{u}^{t'}_i,\bm{u}^{t'}_l; \bm{W}_{\textrm{yy}}), \\
    {\bm{u}}^{t+1}_i = \bm{m}^{t'}_i + U_{\textrm{mlp}}(\bm{m}^{t'}_i; \bm{W}_{\textrm{yy}}).
\end{gather}


We can also incorporate prior knowledge for label-to-label message passing by simple pre-processing: starting from a complete label graph where every node is connected to every other node, we can remove the edge between a pair of nodes if they never belong to any sample simultaneously in the training set.
Depending on a dataset's domain, additional expert knowledge from the domain can be enforced on a dataset-to-dataset basis.
For instance, for predicting species distribution, one can incorporate evolutionary relationships between species, often represented as tree-structures~\cite{ovaskainen2017make,letunic2007interactive}.



\begin{figure}
  \begin{center}
    \includegraphics[width=0.25\textwidth]{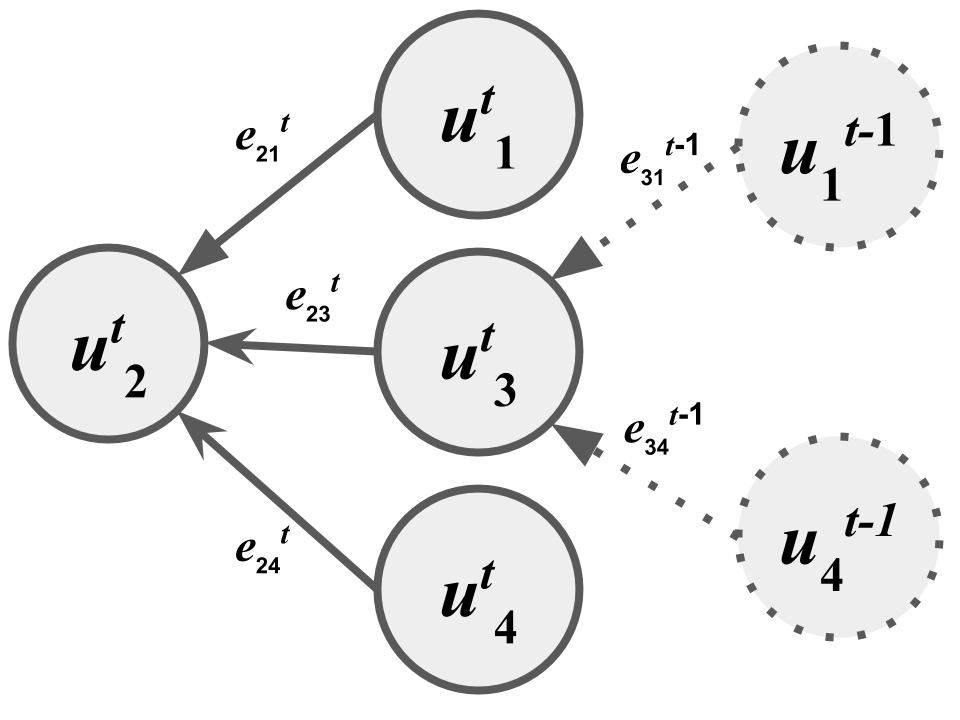}
  \end{center}
  \caption{An example of higher-order correlation between labels. $u_i^t$ is the hidden representation of $y_i$ at time $t$. The figure shows how correlation is formed within the label triplets $\{ y_2, y_3, y_1 \}$ and $\{ y_2, y_3, y_4 \}$.}
  \label{fig:highorder}
\end{figure}

Finally, we describe how HOT-VAE learns higher-order correlation between labels.
If there is a single layer in the decoder, then messages are passed once from labels to labels, which computes correlation between any pair of labels.
Going to higher order, we only need to increase the number of layers in the decoder.
$n$ times of message passing between labels enable learning $(n+1)$-order label correlation.
In Figure~\ref{fig:highorder}, we present a visual explanation, showing how correlation is learned for two label triplets $\{ \bm{y}_2, \bm{y}_3, \bm{y}_1 \}$ and $\{ \bm{y}_2, \bm{y}_3, \bm{y}_4 \}$.
In this example, at time $t-1$, label 3 collects information from labels 1, 4.
At time $t$, label 2 further collects information from label 3.
By this time, two paths has been built from $1\rightarrow3\rightarrow2$ and $4\rightarrow3\rightarrow2$.
Therefore, the presence of label 2 conditions on the label pairs (1, 3) and (4, 3).
This can be easily extended to high orders.

\subsection{Loss Function}
The whole model can be trained in an end-to-end fashion with the Adam optimizer~\cite{DBLP:journals/corr/KingmaB14}.
The overall loss function consists of four parts.
We denote the true binary label vector by $\bm{y}$.
For both the feature and the label branches, we compute binary cross entropy (BCE) over all outputs $\bm{y}_i$ for every sample:
\begin{equation}
    \mathcal{L}_{\textrm{BCE}} = (BCE(\bm{y},\hat{\bm{y}}_f) + BCE(\bm{y},\hat{\bm{y}}_l))
\end{equation}
where BCE is defined by:
\begin{equation*}
    \label{eq:bce_loss}
    BCE(\bm{y},\hat{\bm{y}}) = \frac{1}{L}\sum_{i=1}^{L} -(y_i\log(\hat{y}_i)+(1-y_i)\log(1-\hat{y}_i)) 
\end{equation*}

Further, since the decoder iteratively updates the label node from $t=0$ to $n$, we can also pass those intermediate states from $t=1$ to $n-1$ through a readout layer and enforce BCE loss on these states.
\begin{equation}
    \mathcal{L}_{\textrm{INT}} = (\sum_{t=1}^{n-1} BCE(\bm{y},\hat{\bm{y}}_f^t) + \sum_{t=1}^{n-1} BCE(\bm{y},\hat{\bm{y}}_l^t))
\end{equation}

We also include ranking loss~\cite{6471714} defined as follows, which is shown beneficial in many multi-label tasks:

\begin{equation*}
    RL(\bm{y},\hat{\bm{y}}) = \frac{1}{|{Y}||\bar{{Y}}|} \sum_{(r,s) \in (Y, \bar{Y})} \textrm{exp}(-\hat{y}_r-\hat{y}_{s})
\end{equation*}

where $Y$ is the set of indices for true positive labels and $\bar{Y}$ is the set of indices for true negative labels. $\hat{y}_r$ and $\hat{y}_s$ are the corresponding $r$-th and $s$-th probabilities outputted by the model.
The ranking loss penalizes when a relevant label to the sample is ranked higher than an irrelevant label.
Again, the ranking loss is calculated for both two branches:

\begin{equation}
    \mathcal{L}_{\textrm{RANK}} = (RL(\bm{y},\hat{\bm{y}}_f) + RL(\bm{y},\hat{\bm{y}}_l))
\end{equation}

Finally, with the KL divergence computed for the two branches included, the overall loss function becomes:
\begin{equation}
    \mathcal{L} = \lambda_0 \mathcal{L}_{\textrm{BCE}} + \lambda_1 \mathcal{L}_{\textrm{INT}} + \lambda_2 \mathcal{L}_{\textrm{RANK}} +\beta \mathcal{L}_{\textrm{KL}}
\end{equation}
$\lambda_0, \lambda_1, \lambda_2$ and $\beta$ controls the weights of the four loss terms.

%% file: experiment.tex
\section{Experiments}
We illustrate the power of HOT-VAE on eight real-world datasets covering a variety of application domains including ecology, images, texts, etc.
We first present the main experiment, where HOT-VAE is compared to several other state-of-the-art MLC methods on all the datasets.
We evaluate their performance with conventional metrics such as F-measure and accuracy.
To verify that our model also makes a meaningful improvement in ecology, we perform analysis on 12 metrics measuring discrimination power, calibration, etc. for levels of species occurrence, species richness, and community composition~\cite{norberg2019comprehensive}.
Finally, we present ablation studies showing the benefits of having high-order label correlation and incorporating prior knowledge.

\subsection{Setup}
\paragraph{Datasets.}
The datasets we use to run experiments are: \emph{eBird}~\cite{chen2017deep}, a crowd-sourced bird presence-absence dataset collected from birders' observations; \emph{bibtex} and \emph{bookmarks}~\cite{katakis2008multilabel}, collections of text objects associated with tags; \emph{mirflickr}~\cite{huiskes08} and \emph{scene}~\cite{boutell2004learning}, collections of images with tags; \emph{reuters}~\cite{lewis2004rcv1}, natural language texts with predefined categories based on their content; \emph{sider}~\cite{kuhn2016sider}, side effects of drug molecules; and \emph{yeast}~\cite{nakai1992knowledge}, a biology database of the protein localization sites.
They are all available online~\footnote{http://mulan.sourceforge.net/datasets-mlc.html}~\footnote{https://ebird.org/home}.

These datasets vary in many aspects including the number of samples ranging from 1427 to 87856, number of labels from 6 to 208, feature dimensions from 15 to 368998.
They also cover a wide spectrum of input types: some are raw English text with words ordered sequentially, some are binary features, and the other are real-value vectors (e.g., images).
We present label statistics for each dataset in Table~\ref{tab:dataset_label_stats}, which is useful information to consider when designing models.
One can see that it is common for samples to have more than two labels.
For instance, the median number of labels per sample in $\emph{eBird}$ is 18, which suggests incorporating high-order label correlation into a model will lead to a stronger modelling capacity to learn better joint representations.

We split each dataset into a training set, a validation set, and a test set in the same way as \citet{bai2020disentangled} and \citet{lanchantin2019neural} do.

\begin{small}
\begin{table}
\scriptsize
\begin{minipage}[h]{.47\textwidth}
\centering
\scalebox{0.75}{
\begin{tabular}{c|r|r|r|r|r|r|r} 
 &\#labels  & \multicolumn{1}{l|}{\begin{tabular}[c]{@{}l@{}}Mean\\ Labels\\ /Sample\end{tabular}} & \begin{tabular}[c]{@{}l@{}}Median \\ Labels\\ /Sample\end{tabular} & \begin{tabular}[c]{@{}l@{}}Max\\ Labels\\ /Sample\end{tabular} & \begin{tabular}[c]{@{}l@{}}Mean\\ Samples\\ /Label\end{tabular} & \begin{tabular}[c]{@{}l@{}}Median\\ Samples\\ /Label\end{tabular} & \begin{tabular}[c]{@{}l@{}}Max\\ Samples\\/Label\end{tabular} \\ \hline
\textit{eBird} & 100 & 20.69 & 18 & 96 & 8322.95 & 5793 & 29340\\
\textit{bookmarks} & 208 & 2.03& 1& 44 & 584.67 & 381 & 4642 \\
\textit{bibtex}  & 159 & 2.38& 2& 28 & 72.79 & 54 & 689 \\
\textit{mirflickr} & 38 & 4.80 & 5 & 17 & 1247.34 & 799 & 4120\\
\textit{reuters} & 90 & 1.23& 1& 15 & 106.50 & 18 & 2877 \\
\textit{scene} & 6 & 1.07 & 1 & 3 & 170.83 & 168 & 903\\
\textit{sider} & 27 & 15.3 & 16 & 26 & 731.07 & 851 & 1185 \\
\textit{yeast} & 14 & 4.24 & 4 & 11 & 363.14 & 334 & 903
\end{tabular}
}
\end{minipage}
\caption{Dataset Label Statistics. This shows that many samples have more than two labels to be predicted, and each dataset has a varying degree of label density.}
\label{tab:dataset_label_stats}
\end{table}
\end{small}

\paragraph{Baseline Comparisons.}
We compare HOT-VAE with five other state-of-the-art MLC methods.
MLKNN~\cite{zhang2007ml} is a statisical method based on the k-nearest neighbor algorithm.
SLEEC~\cite{NIPS2015_5969} makes no low-rank assumption, and it learns embbeddings perserving pairwise distances between only the nearest label vectors.
C2AE~\cite{yeh2017learning} is a two-branch autoencoder.
It first produces a latent vector for features and a latent vector for labels; then these two latent embedding are associated by deep canonical correlation analysis (DCCA).
seq2seq~\cite{nam2017maximizing} applies a recurrent-neural-neural (RNN) based encoder-to-decoder model where the encoder RNN encodes features and the decoder predicts each positive label sequentially. 
LaMP~\cite{lanchantin2019neural} consists of multiple attention-based neural message passing modules including one sending messages bewteen features, one from features to labels, and one between labels. 
MPVAE~\cite{bai2020disentangled} is a two-branch disentangled VAE building on a covariance-aware multivariate probit model, which can learn pairwise label correlation.

\begin{small}
\begin{table}
\begin{minipage}[h]{.47\textwidth}
\scalebox{0.75}{
\begin{tabular}{@{}c@{\hspace{0.45em}}c@{\hspace{0.45em}}c@{\hspace{0.45em}}c@{\hspace{0.45em}}c@{\hspace{0.45em}}c@{\hspace{0.45em}}c@{\hspace{0.45em}}|c@{}}
Dataset & MLKNN & SLEEC & C2AE & seq2seq & LaMP & MPVAE & ours\\
\hline
\textit{eBird}    & 0.5103 & 0.2578 & 0.5007 & 0.4768 & 0.4768 & 0.5511 & \textbf{0.5747}\\
\textit{bookmarks} & - & - & - & 0.3620 & 0.3551 & - & \textbf{0.3630}\\
\textit{bibtex}    & 0.1826 & 0.4490 & 0.3346 & 0.3930 & 0.4469 & 0.4534 & \textbf{0.4693}\\
\textit{mirflickr} & 0.3826 & 0.4163 & 0.5011 & 0.4216 & 0.4918 & 0.5138 & \textbf{0.5192}\\
\textit{reuters}   & - & - & - & 0.8944 & 0.9060 & - & \textbf{0.9128}\\
\textit{scene}     & 0.6913 & 0.7184 & 0.6978 & 0.7469 & 0.7279 & 0.7505 & \textbf{0.7762}\\
\textit{sider}     & 0.7382 & 0.5807 & 0.7682 & 0.3560 & 0.7662 & 0.7687 & \textbf{0.7708} \\
\textit{yeast}     & 0.6176 & 0.6426 & 0.6142 & 0.5744 & 0.6242 & 0.6479 & \textbf{0.6498}\\

\hline
\end{tabular}
}
\end{minipage}
~\\

\begin{minipage}[h]{.47\textwidth}
\scalebox{0.75}{
\begin{tabular}{@{}c@{\hspace{0.45em}}c@{\hspace{0.45em}}c@{\hspace{0.45em}}c@{\hspace{0.45em}}c@{\hspace{0.45em}}c@{\hspace{0.45em}}c@{\hspace{0.45em}}|c@{}}
Dataset & MLKNN & SLEEC & C2AE & seq2seq & LaMP & MPVAE & ours\\
\hline
\textit{eBird}     & 0.5573 & 0.4124 & 0.5459 & 0.5260 & 0.5170 & 0.5933 & \textbf{0.6270}\\
\textit{bookmarks} & - & - & - & 0.3290 & 0.3593 & - & \textbf{0.3682} \\
\textit{bibtex}    & 0.1782 & 0.4074 & 0.3884 & 0.3840 & 0.4733 & 0.4800 & \textbf{0.4823} \\
\textit{mirflickr} & 0.4149 & 0.4127 & 0.5448 & 0.4640 & 0.5352 & 0.5516 & \textbf{0.5559}\\
\textit{reuters}   & - & - & - & 0.8575 & 0.8890 & - & \textbf{0.8910}\\
\textit{scene}     & 0.6667 & 0.6993 & 0.7131 & 0.7442 & 0.7156 & 0.7422 & \textbf{0.7567}\\
\textit{sider}     & 0.7718 & 0.6965 & 0.7978 & 0.3890 & 0.7977 & 0.8002& \textbf{0.8026}\\
\textit{yeast}     & 0.6252 & 0.6531 & 0.6258 & 0.5999 & 0.6407 & 0.6554 & \textbf{0.6595}\\
\hline
\end{tabular}
}
\end{minipage}
~\\

\begin{minipage}[h]{.47\textwidth}
\scalebox{0.75}{
\begin{tabular}{@{}c@{\hspace{0.45em}}c@{\hspace{0.45em}}c@{\hspace{0.45em}}c@{\hspace{0.45em}}c@{\hspace{0.45em}}c@{\hspace{0.45em}}c@{\hspace{0.45em}}|c@{}}
Dataset & MLKNN & SLEEC & C2AE & seq2seq & LaMP & MPVAE & ours\\
\hline
\textit{eBird}     & 0.3379 & 0.3625 & 0.4260 & 0.3298 & 0.3806 & 0.4936 & \textbf{0.5350}\\
\textit{bookmarks} & - & - & - & 0.2370 & 0.2939 & - & \textbf{0.2984} \\
\textit{bibtex}    & 0.0727 & 0.2937 & 0.2680 & 0.2820 & 0.3763 & 0.3863 & \textbf{0.3953}\\
\textit{mirflickr} & 0.2660 & 0.3636 & 0.3931 & 0.3333 & 0.3871 & \textbf{0.4217} & 0.4078\\
\textit{reuters}   & - & - & - & 0.4567 & 0.5600 & - & \textbf{0.5748} \\
\textit{scene}     & 0.6932 & 0.6990 & 0.7284 & 0.7490 & 0.7449 & 0.7504 & \textbf{0.7639}\\
\textit{sider}     & 0.6674 & 0.5917 & 0.6674 & 0.2070 & 0.6684 & \textbf{0.6904} & 0.6653\\
\textit{yeast}     & 0.4716 & 0.4251 & 0.4272 & 0.4333 & 0.4802 & 0.4817 & \textbf{0.4885}\\
\hline
\end{tabular}
}
\end{minipage}
\caption{Top: ebF1 scores; Middle: miF1 scores; Bottom: maF1 scores produced by all the methods for each dataset. We mark the best scores with bold texts.}
\label{tab:f1s}
\end{table}
\end{small}

\paragraph{Other details.}
We choose two encoders: one is a three-layer MLP which is used for \emph{eBird}, \emph{scene}, \emph{sider}, \emph{sider}, and \emph{yeast}; the other one is a two-layer FMP from \cite{lanchantin2019neural} for the rest of the datasets.
For label graphs, we use both complete graphs and prior graphs (with edges removed if two labels never correspond to any training sample) for each dataset and select the one exhibiting better performance.
Hyperparameter selections and other training details are in the supplementary material.

\subsection{Main Experiments}
\begin{small}
\begin{table}[h]
\scalebox{0.75}{
\begin{tabular}{@{}c@{\hspace{0.45em}}c@{\hspace{0.45em}}c@{\hspace{0.45em}}c@{\hspace{0.45em}}c@{\hspace{0.45em}}c@{\hspace{0.45em}}c@{\hspace{0.45em}}|c@{}}
Dataset            & MLKNN  & SLEEC & C2AE   & seq2seq & LaMP            & MPVAE & ours          \\
\hline
\textit{eBird}     & 0.8273 & 0.8156 & 0.7712 & 0.8236 & 0.8113 & 0.8286 &\textbf{0.8455}\\
\textit{bookmarks} & - & - & - & 0.9900 & \textbf{0.9917} & - & \textbf{0.9917}        \\
\textit{bibtex}    & 0.9853 & 0.9818 & 0.9867 & 0.9850 & 0.9876 & 0.9875 & \textbf{0.9878}        \\
\textit{mirflickr} & 0.8767 & 0.8698 & 0.8973 & 0.8839 & 0.8969          & 0.8978 & \textbf{0.8980}\\
\textit{reuters}   & - & - & - & 0.9962 & 0.9970 & - & \textbf{0.9971}          \\
\textit{scene}     & 0.8633 & 0.8937 & 0.8934 & \textbf{0.9456} & 0.9025          & 0.9094 & 0.9155\\
\textit{sider}     & 0.7146 & 0.6750 & 0.7487 & 0.5930 & 0.7510          & 0.7547 & \textbf{0.7555}\\
\textit{yeast}     & 0.7835 & 0.7824 & 0.7635 & \textbf{0.8177} & 0.7857          & 0.7920 & 0.7947\\

\hline
\end{tabular}
}
\caption{HAs for all the methods on each dataset.}
\label{tab:ha}
\end{table}
\end{small}

\begin{table*}[t!]
\centering
\resizebox{\textwidth}{!}{%
\begin{tabular}{c|cccc|cccc|cccc}
        & \multicolumn{4}{c|}{Occurrences}                                           & \multicolumn{4}{c|}{Richness}                                           & \multicolumn{4}{c}{Community}                                                                                                                 \\ \cline{2-13}
        & Accuracy        & Discrimination  & Calibration     & Precision       & Accuracy         & Discrimination  & Calibration     & Precision       & Accuracy                          & Discrimination                    & Calibration                       & Precision                         \\ \hline
MPVAE   & 0.3093          & 0.7803          & 679.8019        & 0.3772          & 18.3679          & 0.4425          & 0.4198          & 4.0185          & (0.2208, 0.2443, 0.2318)          & (0.4053, 0.1910, 0.0502)          & (0.3067, 0.3340, 0.3367)          & \textbf{(0.0708, 0.0842, 0.0373)} \\
HOT-VAE & \textbf{0.2062} & \textbf{0.8278} & \textbf{124.96} & \textbf{0.2612} & \textbf{10.8392} & \textbf{0.6279} & \textbf{0.2679} & \textbf{3.0248} & \textbf{(0.1512, 0.2268, 0.1727)} & \textbf{(0.5549, 0.4326, 0.3509)} & \textbf{(0.2033, 0.2233, 0.2633)} & (0.0831, 0.1214, 0.0668)         
\end{tabular}%
}
\caption{HOT-VAE vs. MPVAE on 12 ecological metrics. For accuracy, calibration, and precision, scores are the smaller, the better; for discrimination, scores are the larger, the better. The better scores are marked with bold texts.}
\label{tab:eco}
\end{table*}

In Table~\ref{tab:f1s}, we present the main experimental result evaluated on three F1-scores: example-based F1 (ebF1), micro-averaged F1 (miF1), and macro-averaged F1 (maF1).
F1-score is the harmonic mean of precision and recall of the predictions.
ebF1 is the average of the F1-score for each test sample.
miF1 aggregates total true positives, false negatives, and false positives for all class labels, and computes a F1-score.
maF1 computes the F1-score independently for each class and returns the average with equal weights for all classes.
Larger F1-scores indicate better performance, with the highest possible value being 1, meaning perfect precision and recall.
We note that high ebF1s show strong results over all test samples, high miF1s indicate strong performance on the most frequent labels, and high maF1 implies strong performance on less frequent labels.
We select the model performing best on validation set based on maF1.

We note that \emph{reuters} deals with sequential input and \emph{bookmarks} has over 380 thousand features; hence, we only compare to seq2seq and LaMP, which can handle position information or inputs with extremely high dimensions.
HOT-VAE outperforms all the baseline approaches on ebF1, yielding a 2.13\% improvement to MPVAE and a 5.69\% improvement to LaMP on average.
Our model again produces the best performance within all the methods on miF1, improving MPVAE by 1.63\% and LaMP by 4.88\% on average.
HOT-VAE performs less well for predicting the rare labels on \emph{sider} and \emph{mirflickr}, indicated by maF1, but it still produces an improvement on most of the datasets.

We also test HOT-VAE on Hamming accuracy (HA), which measures how many labels are correctly predicted in all the labels regardless of being positive/negative.
Table~\ref{tab:ha} summaries the HAs of all the methods on each dataset.
seq2seq performs better than HOT-VAE on \emph{scene} and \emph{yeast}, because they directly maximize subset accuracy, and this works well when the number of labels is small.
Otherwise, HOT-VAE has the best HAs.

\subsection{Measuring Predictive Performance on Species}

Ecologists use joint species distribution models for a number of distinct applications.
To determine if HOT-VAE produces ecologically meaningful improvements in model performance we evaluated the 12 metrics presented in Norberg et al. (2019) used to assess performance predicting species occurrences, species richness, and community composition.
The occurrence metrics measure a model’s performance for predicting presence/absence of individual species, the richness metrics measure the ability to predict the total number of species that occur at a given location and time, and the community metrics measure the ability to predict species occurrences at location pairs.
Four metrics are applied for predicting each of species occurrences, species richness, and community composition: accuracy, calibration, precision, and discriminative power of predictions. We compare our model to MPVAE, the best performing baseline method on \emph{eBird}.

The experimental results for the 12 measures are shown in Table~\ref{tab:eco}.
For accuracy, calibration, and precision, smaller values indicate better predictive performance; for discrimination power, larger values are better (calculations of these values are in supplementary material).
HOT-VAE outperforms MPVAE for 11 of the 12 metrics.
We only have worse precision at the community composition level; this is because this community composition metric only considers pair-wise co-occurrence of species, and MPVAE is optimized for modeling interactions specifically between pairs of species.

\subsection{Ablation Studies}
\begin{figure}[t!]
    \centering
    \includegraphics[width=0.45\textwidth]{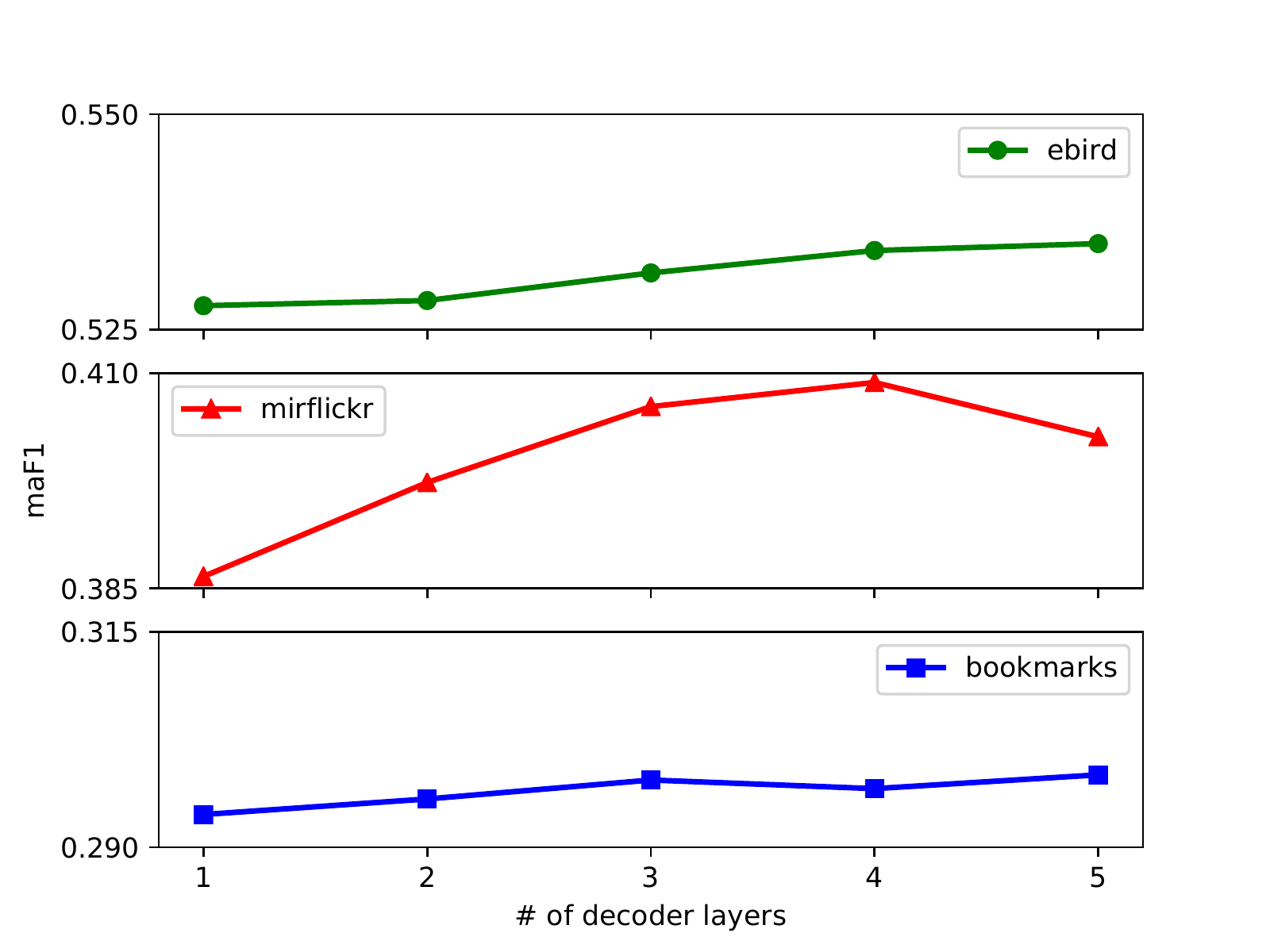}
    \caption{maF1 scores for the datasets \emph{ebird}, \emph{mirflickr}, \emph{bookmarks} when the number of decoder layers increases.}
    \label{fig:ndec}
\end{figure}
We further investigate how incorporating high-order label correlation impacts predictive performance.
We postulate that, for learning $n$-order label correlation, increasing $n$ would be particularly beneficial for improving predictions of rare labels: when treating classifying each label as individual problem, there is very limited information to learn from for the rare labels; including higher-order interactions between labels provides much additional information to a model.

To verify this hypothesis, we train HOT-VAE with different numbers of decoder layers and see how maF1 scores vary under these settings, as high maF1 scores indicate strong results on less frequent labels.
We select three datasets \emph{ebird}, \emph{mirflickr}, and \emph{bookmarks}, which have median labels 18, 5, and 1, respectively.
In Figure~\ref{fig:ndec}, we show maF1 scores for these datasets at $n$ being 1-5.
For the datasets with more labels per sample, larger $n$ has more positive impacts on maF1 scores.
It is also possible increasing $n$ beyond some threshold negatively affects performance as the model may start to overfit.
In general, for the datasets with dense labels, having $n=4,5$ produces best results.
After that, we are likely to hit diminishing returns. 

\begin{small}
\begin{table}[t!]
\centering
\scalebox{0.6}{
\begin{tabular}{c|cccccccc}
         & \emph{mirflickr}       & \emph{ebird}           & \emph{scene}           & \emph{yeast}           & \emph{reuters}         & \emph{sider}           & \emph{bookmarks}       & \emph{bibtext}         \\ \hline
prior    & \textbf{0.8169} & \textbf{0.8259} & \textbf{0.9471} & 0.6857          & \textbf{0.9915} & \textbf{0.5151} & 0.9162          & 0.9405          \\
complete & 0.8116          & 0.8251          & 0.9355          & \textbf{0.6869} & 0.9882          & 0.5120          & \textbf{0.9192} & \textbf{0.9457}
\end{tabular}
}
\caption{medianAUC for each dataset when decoder is on a complete label graph and on a prior label graph.}
\label{tab:prior}
\end{table}
\end{small}

Additionally, we evaluate the effect of incorporating prior knowledge.
We train HOT-VAE on both complete label graphs and prior label graphs, in which we remove the edges between the label pairs if they never positively correspond to any training sample.
To have one metric summarizing the overall performance, we use medianAUC (described in the supplementary material), which represents the probability that a random positive sample is ranked higher than a random negative sample.
In Table~\ref{tab:prior}, we show medianAUC for each dataset when HOT-VAE is trained with these two label graphs respectively.
We see that using prior graphs slightly improves predictive performance in many cases and is at least comparable to using complete graphs in all cases.

%% file: conclusion.tex
\section{Conclusion}
In this paper, we propose an attention-based HOT-VAE for multi-label classification to address the complex relations between real-world objects.
HOT-VAE learns high-order correlation between labels conditioned on features; in other words, not only can it captures relations within multiple objects, but also the relations are adaptive to any feature change.
Experimental results show that HOT-VAE improves over the state-of-the-art techniques.

%% file: supp.tex
\section{Supplementary Material}


We provide additional detail on our experimental analysis.
\subsection{Evaluation Metrics}
We use the following metrics in the experimental evaluation.
\paragraph{F1-scores.}
We denote the number of true positives by $tp$, the number of false positives by $fp$, and the number of false negatives by $fn$. An f1-score is defined as follows:
\begin{equation*}
    F_1 = \frac{2tp}{2tp+(fp+fn)}
\end{equation*}

The example-based F1-score calculates F1-scores for all test samples individually and take the average over them: 
\begin{equation*}
    \mathrm{eb}F_1=\frac{1}{N}\sum_{i=1}^{N}\frac{\sum_{k=1}^L 2y_k^i\hat{y}_k^i}{\sum_{k=1}^L y_k^i+\sum_{k=1}^L\hat{y}_k^i}
\end{equation*}
where $N$ is the number of test samples, $y_k^i$ is the $k$-th ground-truth label of test sample $i$ and $\hat{y}_k^i$ is the $k$-th predicted label of test sample $i$.

The micro-averaged F1-score sums up individual true positives, false positives, and false negatives of all predication outcomes, and use them to compute an F1-score: 
\begin{equation*}
    \mathrm{mi}F_1=\frac{\sum_{k=1}^L\sum_{i=1}^{N} 2y_k^i\hat{y}_k^i}{\sum_{k=1}^L\sum_{i=1}^{N}[2y_k^i\hat{y}_k^i+(1-y_k^i)\hat{y}_k^i+y_k^i(1-\hat{y}_k^i)]}
\end{equation*}
The macro-averaged F1-score is the averaged F1-score over all label classes: 
\begin{equation*}
   \mathrm{ma}F_1=\frac{1}{L}\sum_{k=1}^{L}\frac{ \sum_{i=1}^{N}2y_k^i\hat{y}_k^i}{\sum_{i=1}^{N} [2y_k^i\hat{y}_k^i+(1-y_k^i)\hat{y}_k^i+y_k^i(1-\hat{y}_k^i)]} 
\end{equation*}
\paragraph{Area Under the ROC Curve (AUC).}
AUC can be interpreted as, given a sample, how likely it is for a uniformly randomly drawn positive label is ranked higher than a uniformly randomly drawn negative label.
Formally, an Receiver Operating Characteristic (ROC) curve is a plot showing the performance of a classification model at all classification thresholds.
There are two parameters in the graph: true positive rate (TPR) on the y-axis and false positive rate (FPR) on the x-axis; they both range from 0 to 1.
An ROC curve plots TPR versus FPR at different classification thresholds.
AUC is the entire two-dimensional area underneath an ROC curve from $(0,0)$ to $(1,1)$.
Because at a given classification threshold, we always want that TPR is higher than FPR, AUC is the larger the better.
In our experiment, we compute AUC for each label class and find the median AUC.
AUC is calculated using \texttt{sklearn}, an external Python package.

\paragraph{Ecological Metrics.}
To perform a comprehensive ecological evaluation, we follow~\citet{norberg2019comprehensive}, where they propose 12 metrics on three biological levels: species occurrence, species richness, and community composition.
For each of these levels, they measure predictive performance by accuracy, discrimination power, calibration, and precision.

On the species level, accuracy is the absolute difference between predicted occurrence probability and ground-truth occurrence (1/0), averaged over species and locations.
As for discrimination power, AUC values are computed for all species and are then averaged over species.
Calibration is the absolute difference between predicted and ground-truth occurrence in ten probability bins (each including same the number of samples, based on quantiles), which are then averaged over species.
As a measure of precision, they used the square root of the product of the probability of species presence and the probability of species absence, which are then averaged over species and locations.

For richness and community composition, 100 random matrices of 0/1 species occurrences are sampled based on their prediction probabilities.
The probability of a species at a given site is recalculated to be the mean of the corresponding entries from the 100 matrices.

Richness is the number of locations a species occurs.
On the richness level, accuracy is the root mean squared error (RMSE) between prediction and ground-truth richness.
Discrimination power is measured by Spearman rank correlation between prediction and ground-truth richness.
The quantification of calibration is assessed with the relative frequency, $p$, of test values within the corresponding predictive 50\% central interval, and $|p-0.5|$ is reported.
As a measure of precision, they calculate the standard deviation of prediction intervals and average these standard deviations over the sampling units.

For the community composition level, 300 random pairs of the locations are generated.
For these pairs, three measures of pairwise community similarity are computed: the Sørensen-based dissimilarity $\beta_{SOR}$, the Simpson-based dissimilarity $\beta_{SIM}$, and the nestedness-resultant dissimilarity $\beta_{NES}$~\cite{baselga2010partitioning}.
For prediction values, each pairwise community similarity is computed separately for the 100 matrices, which are then averaged over 100.
Ground-truth pairwise community similarities are also computed.
Then, accuracy, discrimination power, calibration, and precision are measured in the same way as how they are done for the richness level.

Again, we note smaller accuracy, calibration, and precision values indicate better performance, while larger discrimination values are better. 

\subsection{Training Detail.}
We run HOT-VAE on one NVIDIA Tesla V100 GPU with 16GB memory.
To obtain best possible performance, we perform grid search on hyperparameters.
Learning rate is chosen from \{2e-4, 3e-4, 7.5e-4\}; $\lambda_0$ is set to 1; $\lambda_1$ is selected from $\{ 0.1, 0.2, 0.3 \}$; $\lambda_2$ is from $\{ 1, 100, 1000\}$; $\beta$ has a value of 1e-5, 1e-4, or $0$ in some rare cases; $d$ is from \{100, 200, 512\}; and the number of decoder layer is from $\{2,3,4,5\}$.
We use dropout after all layers except the last one; the dropout rate is selected from $\{ 0, 0.1, 0.2, 0.5\}$.
We train the model up to 200 epochs to saturate the performance.
To convert the soft predictions into $\{0, 1\}$ values, we use the same thresholds in~\cite{bai2020disentangled} and select the best threshold for each metric.
We will release the source code upon the acceptance of this paper.